\documentclass{article}
\usepackage[ruled]{algorithm2e}
\usepackage{graphicx}

\SetAlFnt{\small}
\SetAlCapFnt{\small}
\SetAlCapNameFnt{\small}
\SetAlCapHSkip{0pt}
\IncMargin{-\parindent}

\usepackage{fancyhdr} 
\begin{document}


\title{Exploring interactive capabilities for home robots via medium fidelity prototyping}
\date{Manuscript drafted: November 15, 2016}
\author{Martin Cooney$^{1}$, \and Sepideh Pashami$^{1}$, \and Yuantao Fan$^{1}$,
\and Anita Sant'Anna$^{1}$, \and Yinrong Ma$^{1}$, \and Tianyi Zhang$^{1}$,
\and Yuwei Zhao$^{1}$, \and Wolfgang Hotze$^{1}$, \and Jeremy Heyne$^{1}$,
\and Cristofer Englund$^{2}$, \and Achim J. Lilienthal$^{3}$,  \and Tom Ziemke$^{4}$ \\ \\ \and 1 Halmstad University \\  2 RISE Viktoria  \\  3 \"{O}rebro University  \\  4 University of Sk\"{o}vde}

\maketitle


\begin{abstract}
In order for autonomous robots to be able to support people's well-being in homes and everyday environments, new interactive capabilities will be required, as exemplified by the soft design used for Disney's recent robot character Baymax in popular fiction. Home robots will be required to be easy to interact with and intelligent--adaptive, fun, unobtrusive and involving little effort to power and maintain--and capable of carrying out useful tasks both on an everyday level and during emergencies.  The current article adopts an exploratory medium fidelity prototyping approach for testing some new robotic capabilities in regard to recognizing people's activities and intentions and behaving in a way which is transparent to people. Results are discussed with the aim of informing next designs.
\end{abstract}

%
%

%
%






\thispagestyle{fancy}

\pagestyle{plain}

\section{Introduction}

The current article reports on several new robotic prototypes built to explore some potentially important challenges for bringing interactive and intelligent robots into home environments.

Figure~\ref{modelFigureLabel} describes some general motivation for the current work. With a growing challenge of insufficient resources to care for increasingly isolated elderly populations in first world countries, and evidence linking loneliness to health problems, robots could help to support well-being of interacting persons by interacting appropriately. To achieve this, people must first be willing to accept robots into their daily lives, which will require robots to be easy to interact with and useful. Ease of interaction extends to all major components of a robot, both in reacting to humans and proactively behaving; and, usefulness covers many tasks, both everyday and during emergencies. Thus a challenge is that there is a need for ideas which could contribute toward forming a solution for improving quality of life for the elderly, but the design space is highly expansive and designing complete robotic solutions can require much cost in time and resources.

Our approach in the current article involved medium fidelity prototyping to quickly test new interactive capabilities. 
Medium fidelity prototyping strikes a balance between obtaining accurate insight into how a complete system would perform (including in some cases how a system will be perceived by interacting persons) and allowing results to be acquired quickly and practically [Engelberg and Seffah 2002]. For simplicity in the current work we use this term in a general sense to describe an approach between low and high fidelity prototyping; thus, we do not mean that all features of a prototype must be mid-level in terms of completeness or lack details like in ``horizontal'' prototyping [Rudd et al.  1996] but also include approaches for combining aspects of different levels of fidelity referred to as ``vertical''   or ``mixed-fidelity'' prototyping [McCurdy et al. 2006]. Some related concepts include the Wizard of Oz technique in Human-Robot Interaction (HRI) in which teleoperation is used to perform challenging tasks which would be challenging for an autonomous robot [Riek 2012], and the slogan, ``Fail often, fail fast, fail cheap'' which advocates trying many possibilities early on in the development process [Lee et al. 2010].

Following such an approach, nine systems were built, as described in Figure~\ref{protoFigureLabel} and Table~\ref{noveltyTable}, and evaluated.
The main contribution of the current article is a description of insights drawn from testing some new capabilities for a home companion robot, intended to facilitate technological acceptance of robots into homes.

Note: parts 1, 2, and 6 relating to energy harvesting, breath sensing, and transparency, are new; the other results have been partially presented at conferences/workshops (3, 4, 7) [Cooney et al. 2012; Cooney and Karlsson 2015; Lundstrom et al. 2016], submitted to journals (3) [Cooney and Sant'Anna 2016], or described as part of student projects at our university (5, 8, 9) [Ma 2016; Hotze 2016; Zhang and Zhao 2016; Heyne 2015]. All parts relate to interactive capabilities; parts 2, 5, 7, 8, and 9 also relate to intelligent capabilities.

\includegraphics[width=\linewidth]{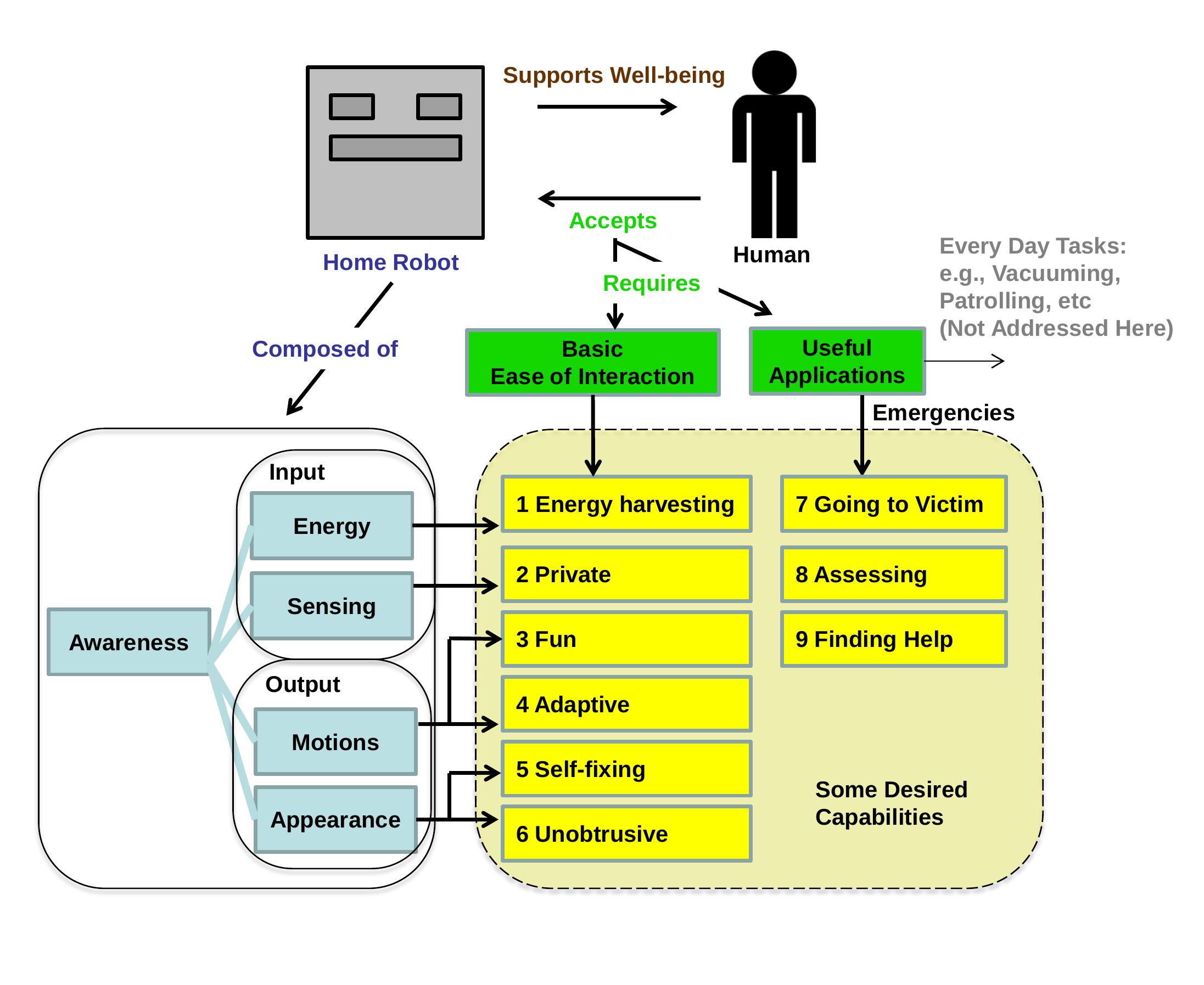}
\begin{figure}[h]
\caption{{\bf A basic goal of home robots is to help people to feel well.
To achieve this, people must be willing to accept robots, which will require robots to be easy to interact with and useful.
Ease of interaction extends to all major components of a robot, enabling proactive and reactive behavior.
Usefulness covers not only tasks such as vacuuming but also helping people in emergencies.
}}
\label{modelFigureLabel}
\end{figure}

\includegraphics[width=\linewidth]{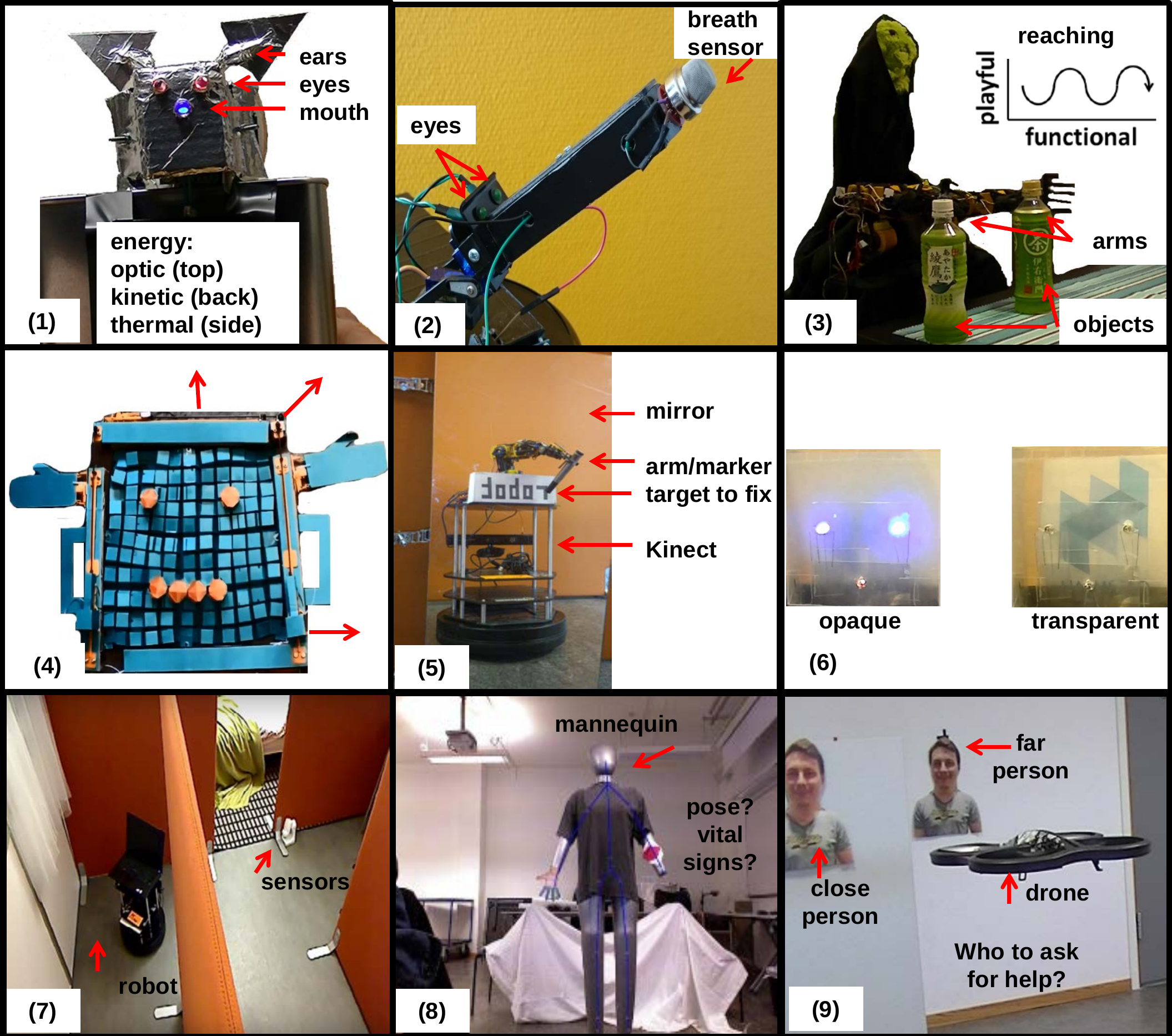}
\begin{figure}[h]
\caption{{\bf Prototypes built: (1) Energy: Energy harvesting, (2) Sensing: Private breath sensing, (3) Motions: Fun reaching, 
(4) Motions: Adaptive size-changing, (5) Appearance: Self-fixing, (6) Appearance: Unobtrusive transparency,
(7) Emergency: Going to a victim, (8) Emergency: Health assessment, (9) Emergency: Finding help.
Note: larger images are provided in each section devoted to a prototype, and a video is also provided with the article.
}}
\label{protoFigureLabel}
\end{figure}

\begin{table}%
\caption{Prototypes described in relation to some previous work.\label{noveltyTable}}{%
\begin{tabular}{|p{0.2\linewidth}|p{0.3\linewidth}|p{0.4\linewidth}|}
\hline
HRI Capability  &  Some past work & Novelty \\\hline

1 Energy: Energy harvesting	
& Robots powered by various environmental energy sources [Kelly et al. 2000; Zivanovic et al. 2009]
& Capability for a robot to provide interactive feedback powered by a person's heat and kinetic energy
\\ \hline

2 Sensing: Private breath sensing 	
& Gas source detection by a mobile robot [Bennetts et al. 2014]
& Reacting to a closeby person via breath sensing
\\ \hline

3 Motions: Fun reaching	
& Some multiobjective motion generation studies exploring the communication of intentions [Dragan et al. 2014; Holladay et al. 2014]
& Design for generating some motions which incorporate fun and functional components [Cooney and Sant'Anna 2016]
\\ \hline

4 Motions: Adaptive size-changing	
& Impressions of short or tall robots [Walters 2009; Rae 2013]; and a design which can be large, tall or wide, but not small [Tachi 2012]
& Some typical impressions of size changes using a design which can be  large, tall, wide, or small [Cooney and Karlsson 2015]
\\ \hline

5 Appearance: Self-fixing	
& Self-detection [Gold and Scassellati 2009]; anomaly-detection [Suzuki et al. 2011]; self-modification [Revzen et al. 2011]
& ``SAS'': combining  these approaches [Ma 2016]
\\ \hline

6 Appearance: Transparent	
& Some methods for achieving transparency such as active camouflaging [Tachi 2003], and colored liquids/soft components [Morin et al. 2014]
& Some typical impressions of transparency using a design combining smart film and conductive plastic
\\ \hline

7 Emergency: Going to a victim	
& Anomaly detection [Novak et al. 2013],  robot navigation [Santos et al. 2013]
& Combining these approaches [Lundstrom et al. 2016]
\\ \hline

8 Emergency: Health Assessment	
& Tele-operated healthcare robots [Katz 2015; Martinic 2014]
& Robotic localization of points of interest for first aid on a fallen person and measuring of some vital signs [Hotze 2016; Zhang and Zhao 2016]
\\ \hline

9 Emergency: Finding help	
& Helpfulness in online reviews [Ghose and Ipeirotis 2011]
& Approaching people estimated to be helpful for first aid [Heyne 2015]
\\ \hline

\end{tabular}}

\end{table}%

\section{Prototype 1 Energy: Energy harvesting}

Home robots require power. Recharging and replacing batteries requires continued attention and effort from humans, robots with docking stations can become stranded, and wireless power transmission is not cost efficient. It would be helpful if robots could seek to secure some power themselves.

A wide range of approaches has been described for how a robot could itself acquire energy, involving wind, water, light, pressure, heat, salinity, and radio waves; some robots have even been designed to mimic animals by acquiring energy from ingesting prey such as slugs and flies, via microbial fuel cells [Kelly et al. 2000; Zivanovic et al. 2009]. We wondered if a social home robot could use energy from a human to interact, but to our knowledge such a prototype had not been designed.

We built a prototype intended to sit on an (elderly) person's lap which secures energy from (1) thermal (the person's body temperature), (2) kinetic (stroking), and (3) optical (light) sources, to provide simple visible interactive feedback, as shown in Figure~\ref{prototype1}. (Light energy, while not directly obtained from a human, can become available when a human turns on a light or takes a robot out of storage to interact.)

\includegraphics[width=\linewidth]{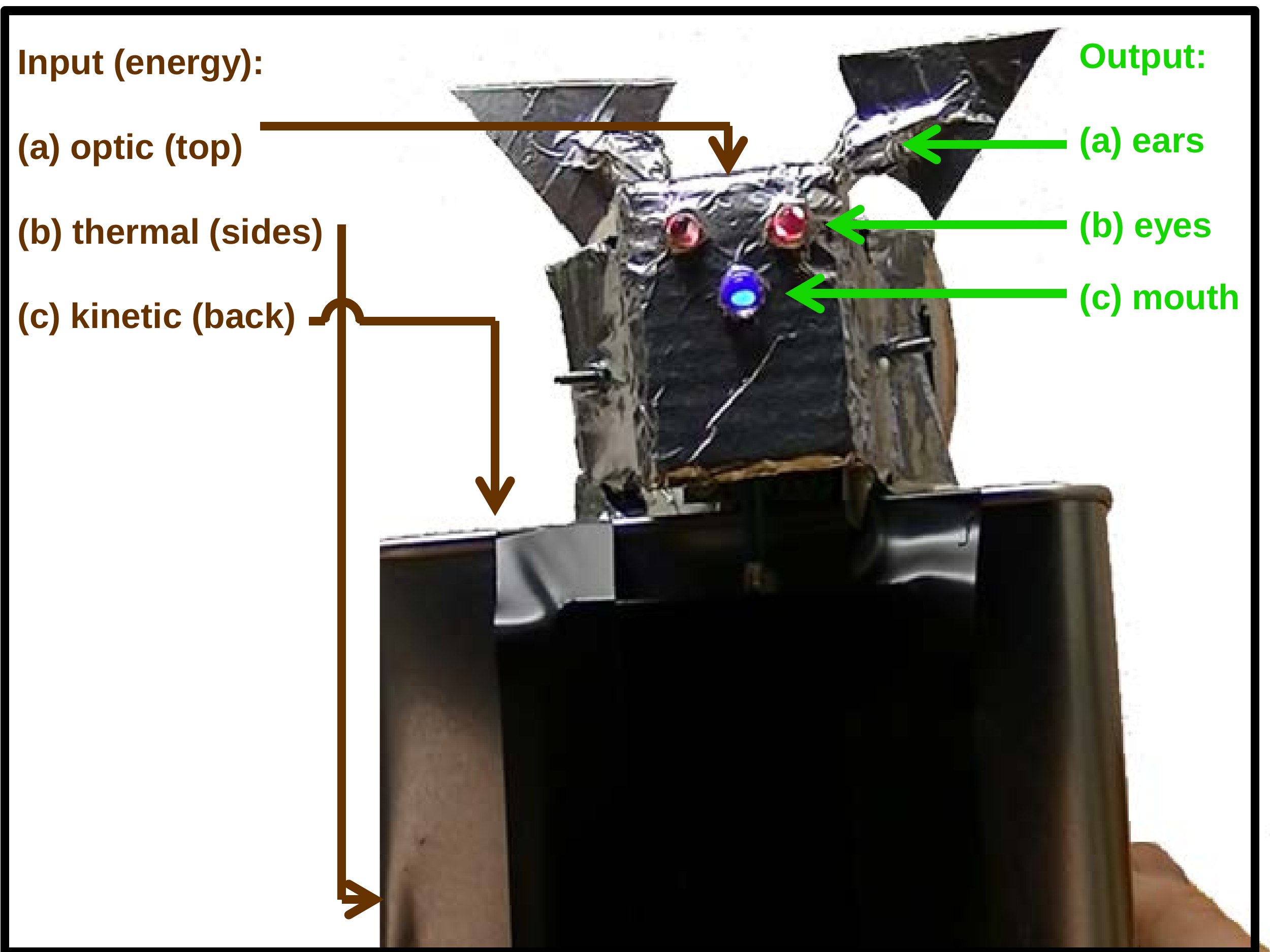}
\begin{figure}[h]
\caption{{\bf Energy harvesting prototype.}}
\label{prototype1}
\end{figure}

\subsection{Approach}

Design was informed by a scenario-specific requirement--the capability to leverage multiple sources of energy simultaneously from a typical holding grasp--as well as general requirements which could facilitate close interaction--human-likeness (to provide a familar interface), light weight and small size. By allowing the palms of a user's hands on the prototype's sides to provide thermal energy, squeezes from the user's fingertips on the prototype's back to provide kinetic energy, and light (e.g., from the sun or a ceiling lamp) on the top of the prototype to provide optical energy, all of the robot's feedback can be seen at once. Easily held (360g; 26.5cm height x 12.3cm width x 8.5cm depth), the prototype's face is a visual focal point in social interactions; therefore the harvested energy is used to power facial parts: Light Emitting Diodes (LEDS) for the eyes and mouth, and two wagging ears.

\subsection{Evaluation}

To objectively compare the prototype in varying conditions, we used OpenCV to measure the magnitude of visual reactions in video recordings of the prototype during interactions. 
For thermal energy, a video was made in which the prototype was held until the LED eyes lit and released five times at a typical illuminance of 500 lux; average time from initiating contact (no red pixels detectable) to full light up (based on counting pixels when fully lit) was calculated using basic image processing to pick colors and remove noise.
For kinetic energy, a video was made while placing known weights on top of the prototype ten times; the minimum pressure to light the mouth higher than a threshold was calculated.
For optical energy, distance travelled by a red marker on one ear was calculated while increasing the amount of light on the energy generator in 100 lux intervals to find an approximate minimum amount of light required to produce a noticeable reaction; distance and frequency at a typical illuminance were also calculated.

\subsection{Results}

For thermal energy, the average time for the LED eyes to light was 7.5 seconds. 
For kinetic energy, LED mouth lighting was visible ((\emph{t}(4) = 3.5, \emph{p} = .02), for a two-tailed t-test) using a 600g weight with 45 cm$^{2}$ area (5.9N, 1300Pa). 
For optical energy, oscillation was observed with a displacement of 0.33cm and frequency of 0.4Hz at 300 lux, and 0.78cm and 0.5Hz at 500 lux.

We feel these results indicate the feasibility of integrating energy harvesting into a small social home robot because: for the thermal reaction, we expect people to generally hold pets or children longer than a few seconds; the touch required to light the mouth was similar to a pat on the back or light hand squeeze [Spears 1985] and much less than the average grip force of a human (approx. 1/34, at 200N [Edgren et al. 2004]); and human environments are often brighter than 300 lux, which is appropriate for large visual tasks with high contrast [IES 1993], and humans are capable of observing smaller radial motion in such conditions [Lappin et al. 2009]. 
We noticed that this ability to operate in typical conditions, along with its simple design, makes the platform easy to demonstrate; we were able to hand out the platform to a visiting class of undergraduate engineering students without worrying about where the demo would take place and without extensive explanation.

\section{Prototype 2 Sensing: Privacy-preserving breath sensing}

In addition to leveraging nearby energy sources, home robots can recognize and react contingently to their sensor input: for example, looking toward an interacting person and tailoring their behavior such that people feel that they matter to the robot. Cameras and microphones can be used for recognition, but such sensors could also be misused to acquire personal data identifying individuals and behaviors, especially in a close physical interaction. Alternatives such as infrared or radar sensors could offer more privacy, but can require placement facing a person or have problems with occlusions and reflections.
One simple solution which could avoid such problems is inspired by the importance of smell in the animal world. Gas sensors, previously used by a mobile robot to detect fires and gas leaks [Bennetts et al. 2014, Lilienthal et al. 2006, Lilienthal and Duckett 2003], present a promising alternative which to our knowledge had not been explored yet for a social robot.
We built a prototype which uses a MQ-135 gas sensor as a breath sensor to recognize an interacting person's relative location in order to react and provide some simple interactive feedback, as shown in Figure~\ref{prototype2}.

\includegraphics[width=\linewidth]{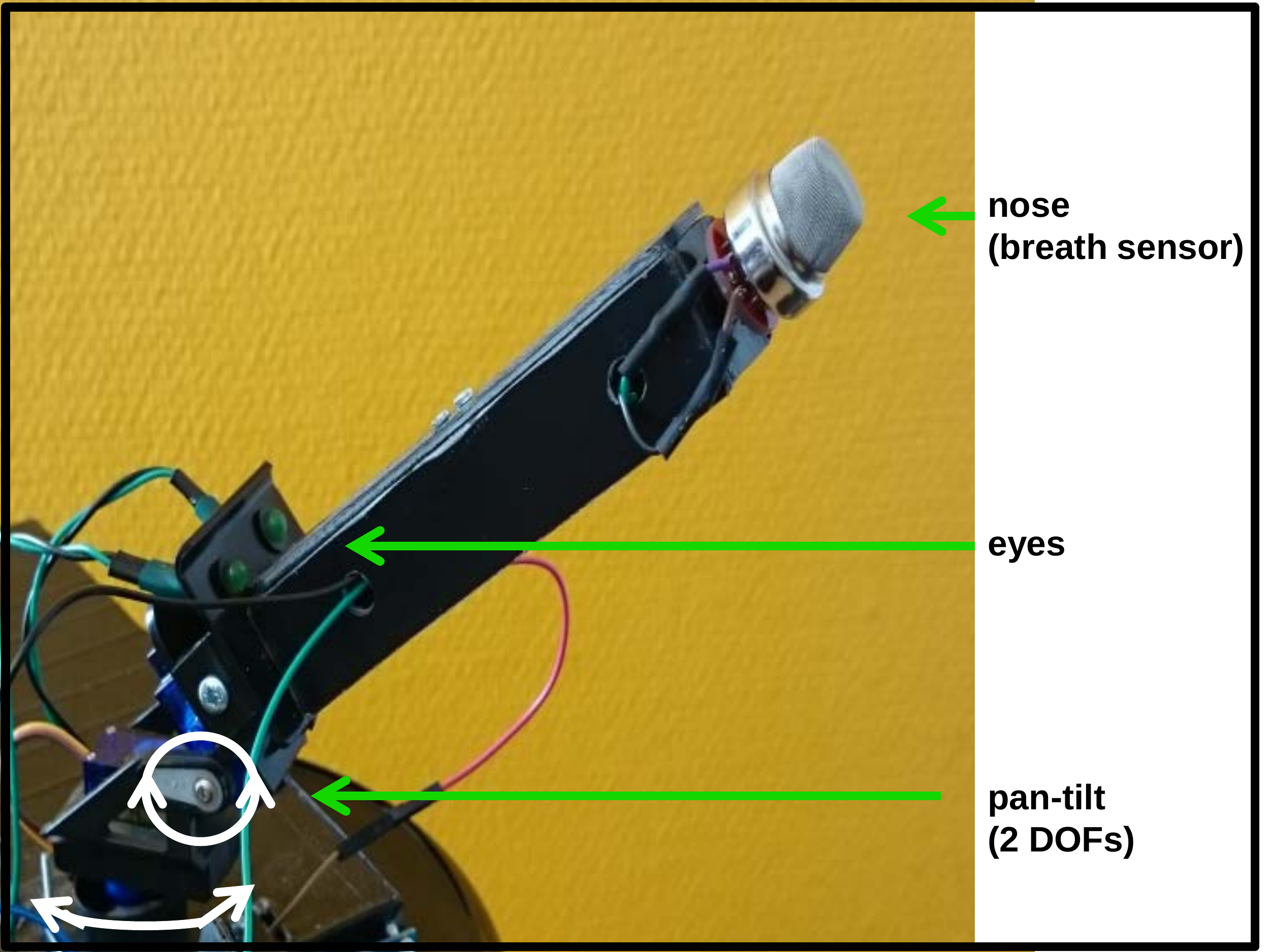}
\begin{figure}[h]
\caption{{\bf Breath sensing prototype.}}
\label{prototype2}
\end{figure}

\subsection{Approach}

Our design took into account scenario-specific requirements--breath sensing capability and a mechanism for conveying an illusion of attention--as well as general requirements which might facilitate interaction--human-likeness  (as a familar interface), lightness and small size. Breath sensing was afforded by a typical inexpensive metal oxide sensor used for air quality control which is also suitable for detecting CO$_{2}$. 

Requirements for the algorithm were obtained from considering some simple desired interactive scenarios: detecting a person's presence or absence in the robot's vicinity, rough location (e.g., left or right), a change in a person's location, and the number of interacting persons (one or two). This suggested that, functionally, the prototype should keep track of its state of belief if there is a human in front of it and orientation, and detect changes. Non-functional requirements were reaction speed and correctness, robustness to noise, and adaptiveness (fixed thresholds to find anomalies, based on recording sensor values when a human isn't nearby, could not be used because sensor values take extremely long to revert to ``normal'' (approx. 1 minute)).
To balance reaction speed and correctness, the robot's algorithm combined a simple but fast approach to recognize sudden large changes and a richer but longer-term approach to recognize slight changes over longer times. 
The simple algorithm uses a state machine with two states (human close or far) and two adaptive thresholds (upper and lower): when a human is close and the sensor value is rising, or the human is far and the sensor value is falling, the simple algorithm adapts its thresholds, sandwiching the current sensor value between new upper and lower thresholds; otherwise when the threshold is crossed the state flips.
The long-term algorithm uses the reweighted norm minimization version of the TREnd Filtering with EXponentials (rTREFEX) algorithm, which models a window of gas sensor output as a series of exponentials; change points are found between the exponentials [Pashami et al. 2014]. 

Additionally, to make use of the recognition capability in an interaction, some simple robot behavior was built with the LEDs and servos. 
Attention was indicated via two degrees of freedom in a pan-tilt configuration to allow the prototype to look in various directions, with a range of approx. 180$^{\circ}$ side to side and approx. 150$^{\circ}$ up and down. Additionally the prototype featured a face with LED ``eyes'' and a nose, and weighed approx. 500g, with a small size of 0.38m height x 0.27m width x 0.27m depth. 
The resulting system was simple but could accomplish tasks difficult for other sensors such as cameras: e.g., detecting a person despite an occlusion, or from the back, or in visually complex environments, or reacting to soft speaking in a noisy environment.

\subsection{Evaluation}

We felt a fundamental question was related to feasibility: could a breath sensor provide value in theory (by preserving privacy) and in practice (by reacting with a reasonable time less than a minute to changes in a person's location)? The answer was unknown because being ``smelled'' could be considered distasteful and intrusive, and precise distance estimation using gas sensors is difficult in general due to air currents and sensor limitations. 

To investigate, data were obtained from seven participants at our university (age: M = 30.1 years, SD = 2.5; 3 female, 4 male). The experimenter asked the participants to read some simple instructions, then started the robot's program on a laptop computer and left the participants alone in a small room. Participants held the prototype on their laps, and five times brought their faces close, spoke to the robot, and then distanced their faces from the robot while pressing the spacebar of the computer to record times. When close, the robot looked up and its eyes lit green; else, the robot looked down and its eyes turned dark. Afterward, participants filled out a simple five-point Likert-style questionnaire with three items asking how comfortable they would feel with the possibility that other people might have access to different kinds of sensors around them and data obtained by these sensors: ``I would feel privacy living in a home with a robot equipped with a'' (camera, microphone, breath sensor). Data acquisition took around five minutes. Impressions of each sensor modality were compared, and average times were calculated.

\subsection{Results}

Impressions of the breath sensor and other sensors (camera, microphone) differed in terms of the degree to which privacy would be perceived: $\chi^2$(1, N = 21) = 6.90, p = .009. Participants strongly agreed that they would perceive privacy in a home with a robot with a breath sensor, somewhat disagreed for a camera, and were neutral about microphones (with camera: 1.9 $\pm$ 1.1, microphone: 3.1 $\pm$ 1.2, breath: 4.6 $\pm$ 0.53). One participant who provided a relatively high score for cameras explained that he does not feel comfortable in front of cameras but that they are everywhere nowadays and he is accustomed to living with them. In the same sense, we think that  differences in perceived privacy might not be observed if the persons with access to the sensor and data are completely trusted and no possibility exists for anyone else to gain access. For timing, the robot reacted to changes in a person's location on average in 6.5 $\pm$ 7.7s (presence: 6.7 $\pm$ 8.7s, absence: 6.3 $\pm$ 6.7s). The positive impression suggested that breath sensors could be useful for a home robot, and we feel the time was acceptable for our context, as we expect people to hold pets for longer than a few seconds.

\section{Prototype 3 Robot motions: Fun reaching} 

In addition to being reactive, home robots will be expected to perform useful tasks in an enjoyable manner, but playful behavior can result in undesired impressions such as that a robot is obnoxious, untrustworthy, dangerous, moving in a meaningless fashion, or boring.
Previous work described dynamic generation of reaching motions intended to appear legible or deceptive [Dragan et al. 2014; Holladay et al. 2014]. Our own work suggested how a similar approach could be used to generate playful motions which avoid typical pitfalls, based on integrating straight useful motions with a curved playful component.
We integrated our model into a motion planning framework to dynamically generate reaching trajectories, built a humanoid robot prototype to perform planned motions , as shown in Figure~\ref{prototype3}, and generated abstracted "point light display" videos from the robot's reaching motions to explore how motions are perceived. (Point light displays consisting of moving dots allow motions to be observed, while hiding the exact details of other factors such as the form, color, and size of a moving artifact [Johansson 1973; Veto et al. 2013]).

\includegraphics[width=\linewidth]{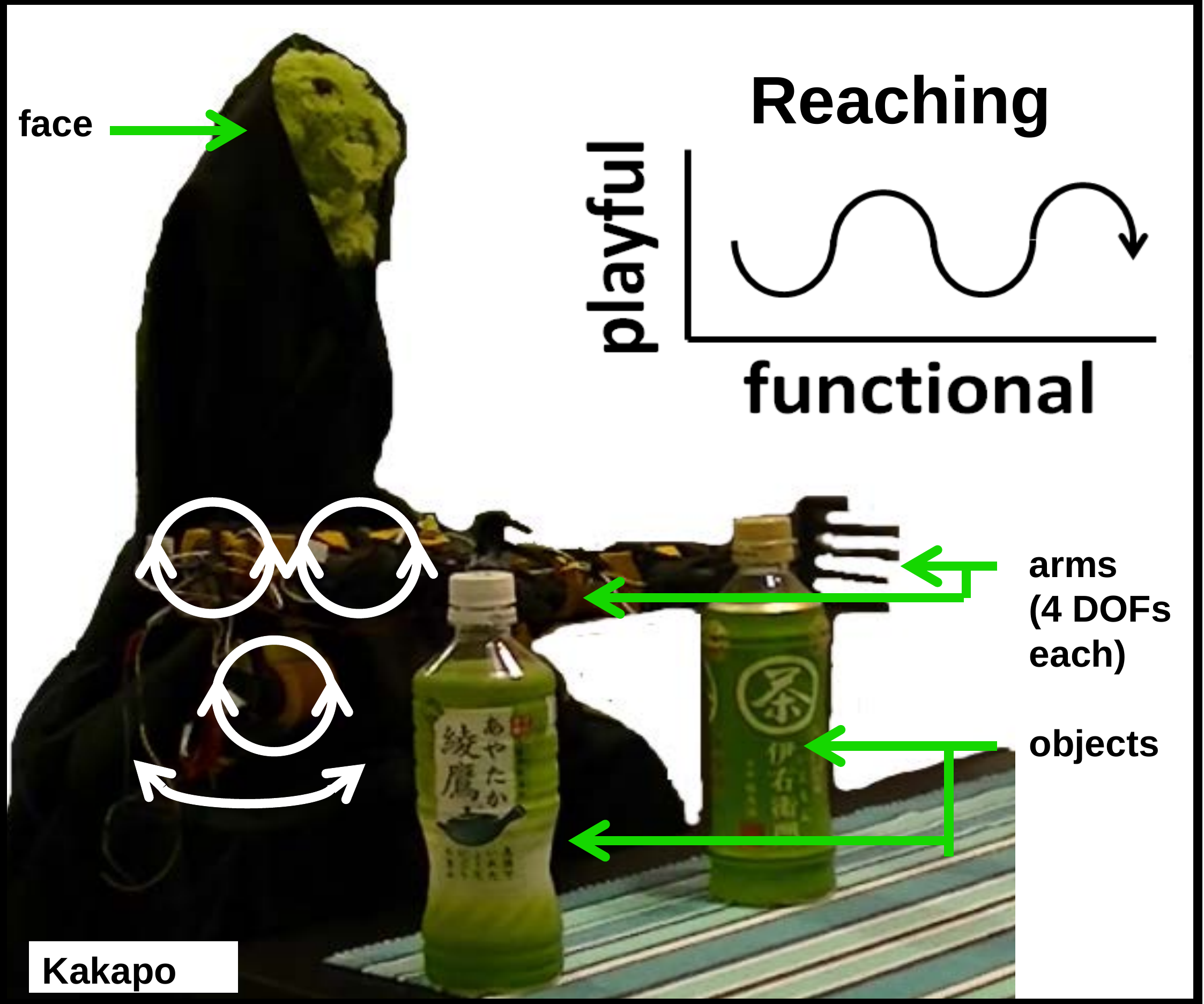}
\begin{figure}[h]
\caption{{\bf Fun motions prototype.}}
\label{prototype3}
\end{figure}

\subsection{Approach}

We built a prototype based on scenario-specific requirements--capability to dynamically generate motions which could be perceived as playful--as well as general requirements which might facilitate interaction--human-likeness  (as a familar interface), and ability to reach objects.
We predicted that failures in playful motions could occur as a result of over-playfulness, hidden goals, wildness, randomness and lack of variety, and if so that they could be avoided by planning motions to be helpful, autotelic (having no goal outside of playfulness), safe, clear in purpose, and anomalous. To dynamically generate motions we adapted a gradient descent framework called Covariant Hamiltonian Optimization for Motion Planning (CHOMP) [Zucker et al. 2013], combining straight reaching motions with a curved playful component. For human-likeness the prototype was given a face and ability to reach objects was implemented by giving the prototype two arms at approximately human arm height and a mobile base.

\subsection{Evaluation}

To investigate how to design successful playful reaching motions, we compared the proposed motions with a baseline (some naive playful motions which did not take into account our guidelines), both for single motions and sequences. Four videos were generated by having the robot reach for an object on a tabletop. To obtain more general impressions LEDs were attached to the arm and objects to create point-light displays. Videos 1 and 3, consisting of single motions and sequences using the proposed model, were designed to provide support that the model is perceived as playful. Videos 2 and 4, consisting of single naive motions and a naive sequence surrounded by proposed sequences, were designed to investigate effects of failures. The videos were watched by 14 participants at our university (five females and nine males; average age = 30.5 years, SD = 10.2 years) over approximately one hour.
To obtain a wide range of feedback, both qualitative and quantitative, we used the think-aloud method, questionnaires, and continuous evaluations. 

\subsection{Results}

The proposed curved motions were perceived as more playful than straight motions (3.6 $\pm$ 0.43 vs. 2.3 $\pm$ 1.1 out of 5: \emph{t}(13) = 4.5, \emph{p} = .001), with scores from both the questionnaires and continuous scoring also above neutral: (5.5 $\pm$ 1.2 vs. 4.0 and 3.6 $\pm$ 0.43 vs. 3.0): \emph{t}(13) = 4.8, \emph{p} $<$ .0005 ( $<$ \emph{$\alpha$} = .02), \emph{t}(13) = 4.9, \emph{p} $<$ .0005, and participants laughed during 20 playful motion sequences.

The model was also perceived as failing significantly less than the baseline for questionnaire scores: \emph{t}(69) = -5.8, \emph{p} $<$ .0005, and think-aloud comments (\emph{$\chi^2$}(1, \emph{N} = 35) = 22.8, \emph{p} $<$ .001). In Video 4 mean goodness scores before and after the baseline sequence differed significantly (with one participant's data removed because he did not rate the robot at all before the baseline motions): 3.2 $\pm$ .87 before vs. 2.3 $\pm$ 1.2 after, \emph{t}(12) = 3.4, \emph{p} = .006, and playfulness was not regarded a failure in the system, with mean perceived goodness before the baseline motions (3.0 $\pm$ 1.0) higher than a score of 2.0 expressing slight disagreement that the robot performs well, \emph{t}(13) = 3.7, \emph{p} = .003. 

Thus, our results suggested that, in the simplified scenario investigated, the prototype could generate some playful motions which are also perceived as good.

\section{Prototype 4 Robot motions: Adaptive size-changing}

In addition to moving in an entertaining way, a robot could also seek to support positive interactions by moving in such a way as to adapt to people's preferences and routines [Dautenhahn et al. 2004]. One way in which a robot could adapt itself is by changing its size: for example, a robot could become large to attract the attention of someone who is deep in thought, or small to be carried by a person who changes location frequently. Such size changes should not be threatening or bothersome.
How people perceive different heights in a robot has been previously investigated [Walters et al. 2009; Rae et al. 2013], but not different widths or size changes. As well, some mechanisms for size changes have been proposed, including a folding structure which can become shorter or thinner but not at the same time [Tachi and Miura 2012]. We built a prototype which can become taller or wider, or both at once, as shown in Figure~\ref{prototype4}., and used it to gather typical impressions of how size changes are perceived.

\includegraphics[width=\linewidth]{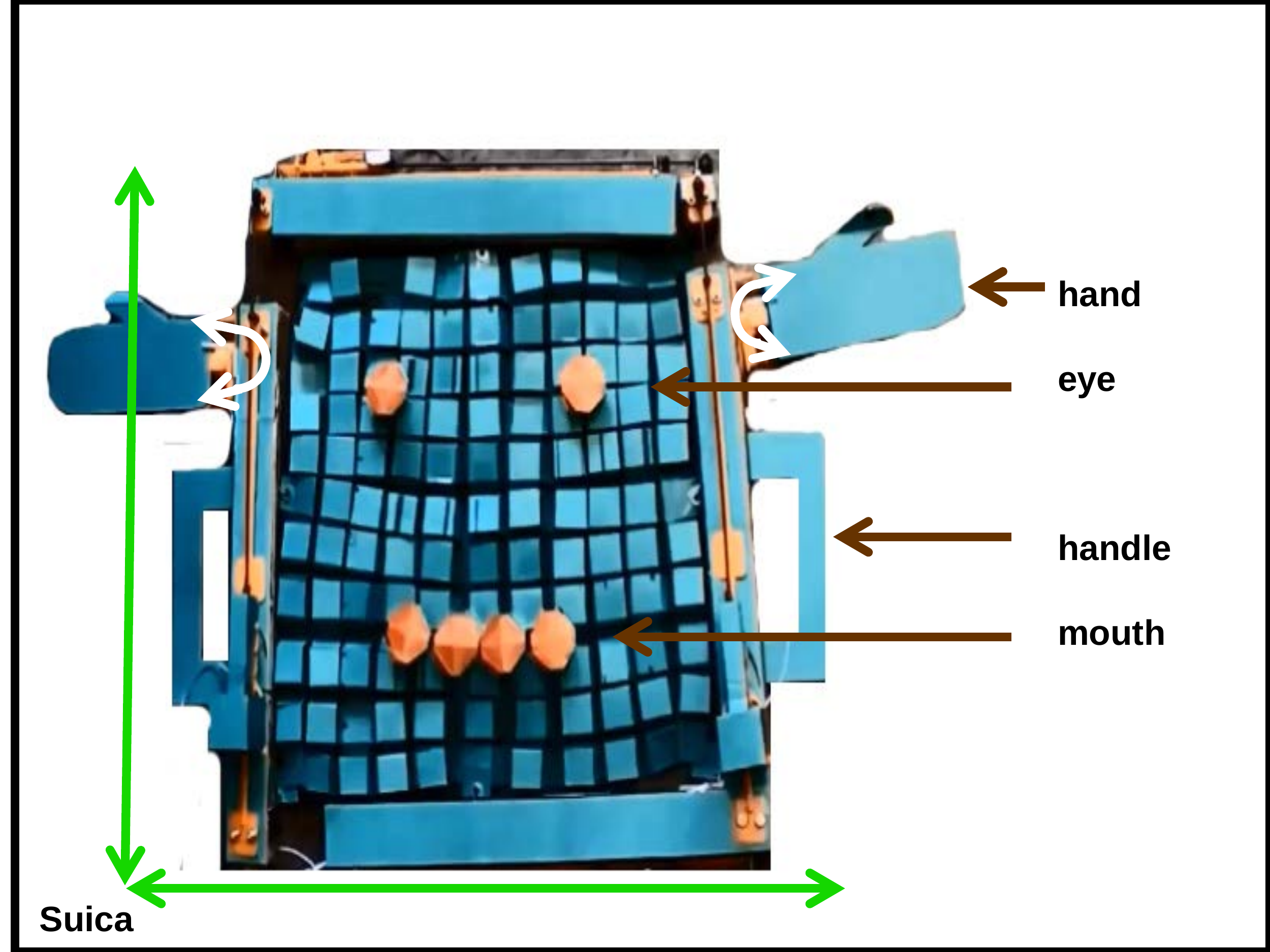}
\begin{figure}[h]
\caption{{\bf Size adapting prototype.}}
\label{prototype4}
\end{figure}

\subsection{Approach}

Our design was based on a scenario-specific requirement--capability to perform size changes in height and width as one cohesive whole--as well as general requirements which might facilitate interaction--human-likeness (as a familar interface), light weight, and small size. Actuation in two dimensions was realized by forming a frame from parallel groups of linear actuators; the challenge was to create a structure inside the frame which could change size independently along two different dimensions while appearing to be a complete artifact, and to do so in a safe and power-efficient manner (which would be difficult with a simple solution involving stretchable rubber). We created such a structure by designing a folding pattern composed of a grid of square components linked by chevron-shaped connectors. Additionally we attached a face and moving hands, a camera, and Bluetooth. The implemented prototype was light (approx. 700 grams), and small but capable of expanding up to eight times in area, and approximately three times along a principal axis (from 27.5cm to 77.5cm). 

\subsection{Evaluation}

We did not know how size changes would be perceived. Expansions could appear intimidating, contractions could show tension or cuteness, and repeated changes could seem playful, but the literature suggested many other possibilities.

To identify typical impressions we asked eight participants (age: M = 33.5 years, SD = 9.6; 2 female, 6 male) to speak their thoughts aloud while watching the prototype change size. Alone in a room with the experimenter participants read a short handout, watched seven size changes in random order (tall, short, wide, thin, large, small, repeated), and provided feedback in short interviews, over approximately thirty minutes. 

Typical impressions expressed by more than one participant were extracted from coded transcripts and analyzed with regard to valence (how positive impressions appeared to be), consistency across similar conditions for expansions and contractions (e.g., comparing impressions for tall and wide), as well as agreement with our expectations. 

\subsection{Results}

Impressions spanned a spectrum from positive to negative; significant differences were not observed in numbers of positive, neutral, and negative impressions: $\chi^2$(2, N = 54) = 0.8, p = .7.

Impressions also exhibited some consistency between related categories: expansions were perceived by some as intimidating/angry (tall and large), or incredulous (tall and wide) due to the expanding body and eyes. Contractions seemed fearful and attentive (all three), and attractive (thin and small). Two unshared impressions for expansions were that widening seemed like the prototype was happy and smiling due to the expansion of the mouth, and that expansions in both width and height seemed unnatural due to dispersion of the face. Unshared impressions for contractions were that the robot was sad or angry (for short) which like the fearful impression might have been perceived as a reason for shrinking away from interaction.

Thus, some impressions were similar to what we had expected but expressed differently (e.g., the intimidating robot was angry, the cute robot was attractive or nice, tension was described as fearfulness or attentiveness, and playfulness was described as happy and excited). Unexpected impressions of incredulousness and happiness also resulted from expansions of the facial components (eyes and mouth). We feel this supported our approach of checking the overall kinds of impressions which would result as our first step, rather than starting with a more specific kind of study (such as a forced-word test).

\section{Prototype 5 Robot appearance: Self-fixing}

Alongside motions, a robot's appearance is also important for safety, trust, liking, and aesthetics (e.g., avoiding uncanny impressions) and should be maintained. For a robot, maintaining appearance is a basic problem because the appearance can be easily accessed without requiring internals to be exposed; furthermore, maintenance is an important problem for home robots because physically embodied systems experience wear and tear and laypersons typically lack the parts and knowledge to conduct repairs. It would be beneficial if robots could help to fix themselves.

Passive fixing with self-healing materials will be useful in the future, although challenges exist in dealing with large, repeated damage [Blaiszik et al. 2010] or cases when material should be removed or aligned (e.g., hair or fur). Active fixing, in which a robot moves to fix itself, can address such cases but requires capability for self-detection [Gold and Scassellati 2009], anomaly discovery [Suzuki et al. 2011], as well as self-modification [Brodbeck et al. 2012; Revzen et al. 2011], which to our knowledge have not been combined previously.

We built an active fixing prototype comprising a mobile base, a target for self-fixing (a small sign with the word, "Robot"), and an arm with a marker, as shown in Figure~\ref{prototype5}.; the prototype (assuming no task has been given by a human) wanders through a home-like environment in search of a reflective surface such as a mirror, checks the appearance of its sign for an anomaly (a missing letter), and writes in the letter using the marker held in its arm.

\includegraphics[width=\linewidth]{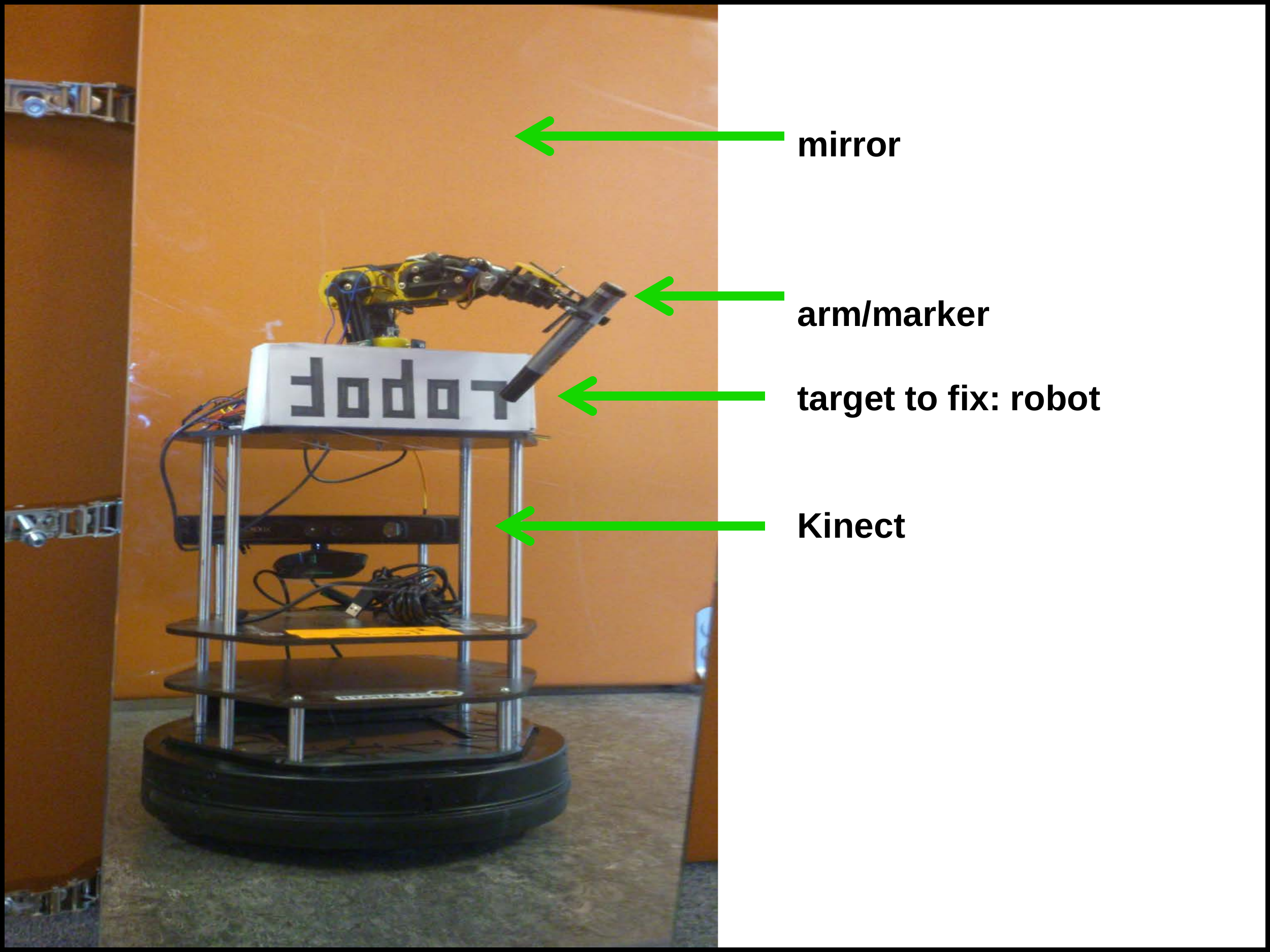}
\begin{figure}[h]
\caption{{\bf Self-fixing prototype.}}
\label{prototype5}
\end{figure}

\subsection{Approach}

Inspired by how humans check their appearances in a mirror, our prototype was designed to perform self-fixing in three steps: self-detection, anomaly detection, and self-modification.
For self-detection the robot navigated to find a mirror and detected itself visually. We used Robot Operating System (ROS) to localize the robot in a known map and calculate paths, and OpenCV's template matching functions along with known images of the robot and some empirically determined thresholds for visual detection.

To recognize anomalies, a one-class classifier was trained on features from various regions on an extracted image of the robot's sign. First the image was divided into five equal regions based on our prior knowledge that there would be five letters, then the number of SIFT features in each grid cell was passed to a one-class Support Vector Machine (SVM) classifier with an RBF kernel and parameters nu and gamma set to 0.5 and 3.1e05, using LIBSVM.

For self-modification we conceived of a simplified model for depicting alphanumerics with six lines or less, inspired by the seven segment approach (https://www.google.com/patents/US1126641), and recorded joint values; by recording two points for each grid cell such as the center and upper right hand corner any alphanumeric could be drawn.

\subsection{Evaluation}

Self-detection was assessed by commanding the prototype 20 times to look for itself from an arbitrary starting point while wandering randomly in a 3m x 3m home-like environment, with a mirror placed in four different areas: the bathroom, bedroom, kitchen, and entrance; the robot halted if it detected a positive (an image it thought represented itself) or visited each room once.
The accuracy of anomaly detection was measured using roughly equal numbers of normal and anomalous images (70 and 61) obtained by placing the robot in different poses relative to the mirror (distance and angle), in different light, and with different degrees of anomaly occurring to the last letter (the "t" of "Robot").
The robot's sign was checked before and after fixing by counting the number of SIFT points detected in the grid cell containing an anomaly.

\subsection{Results}

The prototype detected itself correctly in 76.9\% of the cases in which it detected a positive (10/13), and did not find a mirror seven times; for anomaly detection, accuracy was 71\%; and the number of SIFT features in the anomalous grid cell of the sign increased from almost none to a reasonable amount (3 to 37).
Successful self-detection took on average approx. 150 seconds, whereas anomaly detection and self-modification were quick, requiring only several seconds.

We feel the results are reasonable due to the complexity of the challenge: the self-detection module did not handle differences in scale or rotation (the robot's distance or angle relative to the mirror), the robot could miss itself when it is turning (in between frames), and the complex environment had many objects colored similar to the robot; for anomaly detection there was high variation in pose and illumination; and for self-modification the basic shape of the letter could be seen and the number of SIFT points was similar to results for a normal image, suggesting that fixing resulted in some improvement. Moreover, if we assume the mock-up home was ten times smaller than an average home (in Sweden this is 89m$^2$), the average time taken would be approx. 25 minutes, which would not be prohibitive if the robot occasionally has some down-time (e.g., when humans are sleeping or not at home), locations of previous detections can be remembered, and other tasks can be conducted simultaneously (e.g., looking for potential danger).

\section{Prototype 6 Robot appearance: Unobtrusive transparency}

In addition to reducing maintenance times, robots in human environments can also seek to recognize people's activities and intentions and reduce the degree to which they obstruct or get in the way of humans. One desirable characteristic could be to avoid blocking a human's view, which a robot could realize by turning transparent. 
But, how transparency in a social robot would be perceived was unclear. 
Various mechanisms are being explored for transparency, including active camouflage which requires projectors or many LEDs [Tachi 2003]. Some techniques have also become feasible at the nano-scale but not yet at the scale of robots which could interact with humans [Valentine et al. 2009]. Transparent organic LEDs (OLEDs) could be used, but white backgrounds can be problematic [planar 2016]. Furthermore, liquids of varying opacity can be moved within see-through materials, using various components such as pumps and reservoirs [Morin et al. 2014].
We report on a simple, light design using smart film and conductive plastics, used to explore typical impressions of a prototype becoming transparent (referred to here as  ``\emph{transparification}''), as shown in Figure~\ref{prototype6}.

\includegraphics[width=\linewidth]{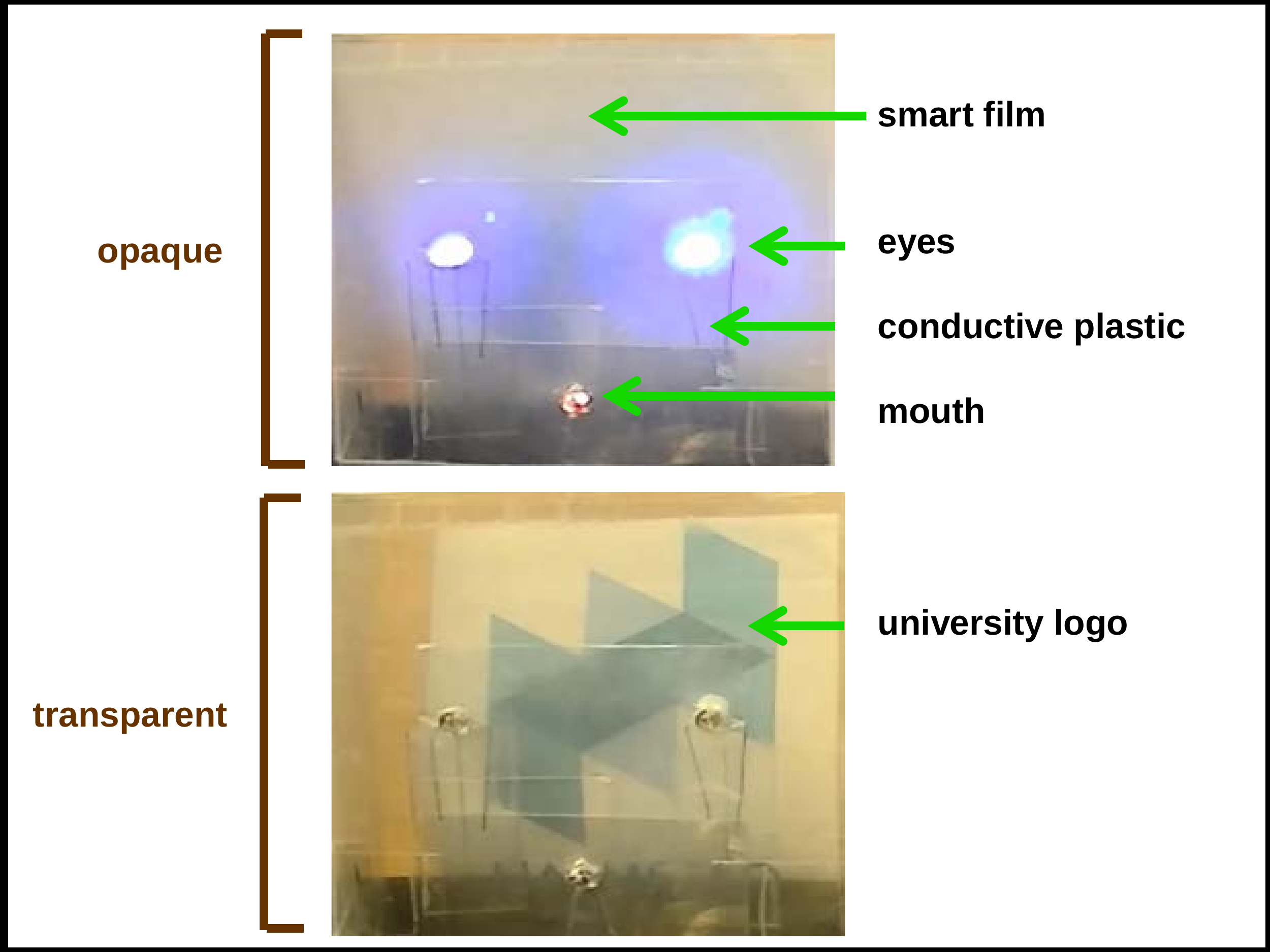}
\begin{figure}[h]
\caption{{\bf Transparency-capable prototype.}}
\label{prototype6}
\end{figure}

\subsection{Approach}

We built a prototype based on scenario-specific requirements--capability to turn transparent or opaque--as well as general requirements which might facilitate interaction--human-likeness, lightness and small size. Transparification was realized by using two electrochromic films containing polymer dispersed liquid crystals (PDLCs) which align to let light pass when powered behind the main parts of the prototype, clear light emitting diodes (LEDs), and polyethylene terephthalate (PET) plastic coated on one side with indium tin oxide (ITO) to conduct electricity to the LEDs and act as touch sensors. The prototype was also given humanoid characteristics (a head and actuated hand, attached above an opaque base holding electronics), weighed approx. 700g and measured 0.35m width x 0.175m height x 0.105m depth.

\subsection{Evaluation}

We did not know what kind of social impressions could result from transparification, because humans cannot turn transparent and many possibilities existed (e.g., clarity and understanding, fear, shadiness, or embarrassment could be attributed). We predicted that proactive transparification would be perceived positively, as an attempt to allow a person to see better, but that as a reaction to a human behavior it would indicate a negative feeling toward interacting. Furthermore repeated transparification would indicate playfulness or desire to be noticed in the proactive case, and acknowledgement of human behavior in the reactive case.

To check, the prototype was shown to eight participants (age: M = 33.4 years, SD = 11.0; 2 female, 6 male), who were asked to describe the robot's behavior aloud (e.g., what is the robot doing, and why do you think the robot did that?). 
Two factors were controlled: the robot behavior (transparification, opacification, or changes repeated three times each at 1Hz) and timing (proactive and reactive). In the reactive case participants were asked to wave, say hello to the robot, and touch its head before the robot's behavior was triggered (imagining that they were interacting with the robot). Transcripts were coded and typical subjective comments common to two or more participants were extracted.

\subsection{Results}

As shown in Table~\ref{transpResultsTable}, transparification was perceived by half of the participants as indicating a change in arousal, with the robot turning off; a sleeping metaphor was common, as well as references to attentiveness, and other emotional impressions related to valence or dominance were not perceived. Repeated changes were described as "blinking", and sometimes as the robot calling for attention or seeking interaction. Proactive behavior was sometimes unclear, whereas the reactive robot sometimes appeared to malfunction.

\begin{table}%
\caption{Typical impressions of transparent behavior.\label{transpResultsTable}}{%
\begin{tabular}{|p{0.1\linewidth}|p{0.25\linewidth}|p{0.25\linewidth}|p{0.25\linewidth}|}
\hline
{\bf } & {\bf Transparent} 	& {\bf Opaque} & {\bf Repeated } 			 	
\\ \hline
Proactive 	
& fell asleep (5), turned off (4), inactive (2)  
& turned on (4), woke up (2), waiting (2), attentive (2), unclear (2)	
& blinking (5), waiting (3),  attentive (3), calling attention (2), unclear (2)
\\ \hline
Reactive 	
& turned off (4), broken (3), responded (2), inactive (2), fell asleep (2) 	
& responded (7), woke up (3)
& blinking (4), responding (4), waiting (2), broken (2), seeking interaction (2) 
\\ \hline

\end{tabular}}

\end{table}%


Reasons were derived from asking participants. Transparification was perceived as “turning off”, like a screen turning dark; we thought this could be due to a reduction of information transmitted from the robot. Anthropomorphic impressions of sleeping or blinking were catalyzed by the prototype's humanlike qualities. Repetition indicated desire; we thought this could be because adaptors (repeated motions in humans) like finger- or foot-tapping can express concern. Reactive behavior seemed clearer due to causality inferred from temporal correlation of behaviors. Some impressions of malfunctioning were due to perceived incongruity in the robot turning off after being greeted by the person.

Thus, the results suggest that transparification can be incorporated into a home robot's capabilities, although care should be given in a reactive context to avoid an impression of malfunctioning; and that one possible use could be to indicate a robot is dormant, e.g., when a person is busy and not interacting.

\section{Prototype 7 Useful application (fall emergency): Smart home integration}

In addition to interacting in a nice way, home robots should be capable of performing useful tasks, one of which will be helping people in emergencies. 
One problem is that people might not want to be normally observed by cameras and microphones on a robot (e.g., when in the bathroom or bedroom), and also a robot might not become aware of an emergency happening in a different part of a home.
Here we propose that a robot can be combined with some environmental sensors, which can be simple to preserve
privacy, and placed throughout a home to detect emergencies when a robot is not nearby. 
The interactive feature proposed is that, when trouble is suspected, the robot can go to where a person is and ask if they are okay.
Verbal communication is a common and expressive interactive modality which is also useful because a robot's positioning does not have to be perfect (conversation can take place over distances without requiring line of sight). Some home robots have been designed to go to speak with a person when a condition is met; e.g., one robot urges elderly persons to drink if they have not had a drink for a while  [Dragone et al. 2015]. As well, some smart home systems have been designed to detect anomalous behavior patterns [Novak et al. 2013; Kim et al. 2010]. Here we combine these two approaches, building a prototype which, based on anomalies detected using some simple environmental sensors, can move close to a victim and ask if they are okay, as shown in Figure~\ref{prototype7}.

\includegraphics[width=\linewidth]{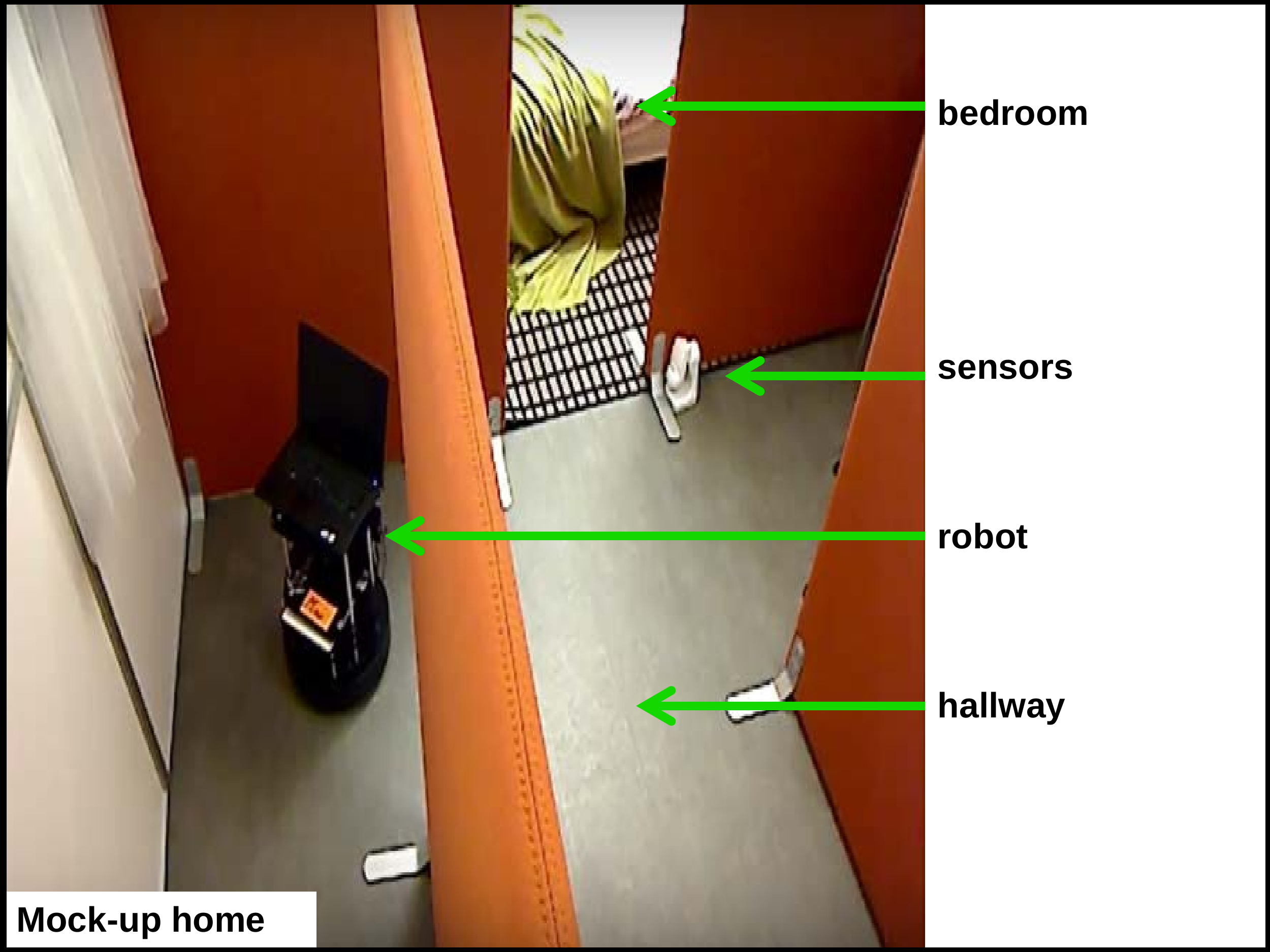}
\begin{figure}[h]
\caption{{\bf Smart home integrated prototype.}}
\label{prototype7}
\end{figure}

\subsection{Approach}
We built a prototype based on scenario-specific requirements--capability to process data from simple environmental sensors in a central database--as well as general requirements which might facilitate interaction--human-likeness and mobility--consisting of simple sensors, a central database, and a mobile robot.
Eleven sensors (four pressure, three contact and four passive infrared) were placed in four locations (a bathroom, bedroom, hallway, and kitchen) in a 3m x 3m small single-floor apartment-like space. These sensors could be useful for example for a person with dementia: pressure sensors can detect if a person falls and stays in one spot for a long time; contact sensors can detect if a person is opening drawers to cook even though they have already eaten; and infrared sensors can detect someone leaving the house in the middle of the night. Data from the sensors was gathered in a central database. A random forest classifier [Ho 1995] was trained to detect anomalous patterns, which triggered the robot to move to the location of the anomaly to investigate.
The prototype was given the capability to communicate in a human-like fashion via speech; it was designed to verbally ask if it should call emergency medical services (EMS); a negative response caused the robot to return to its initial position, and a positive answer or timeout caused the robot to state that a call for help had been made.
For the robot's hardware, a small differential drive mobile base with a Microsoft Kinect sensor were used (0.35 x 0.35 x 0.42m, 6.3 kg). For software, Robot Operating System (ROS) was used for navigation and visualization, and Festival and CMU Pocket Sphinx for speech-based interactive capabilities.

\subsection{Evaluation}

To be feasible, a robot would have to be able to navigate to the scene of an anomaly quickly and correctly determine if an emergency had occurred through verbal interaction. Evaluation was conducted by having the experimenter trigger the environmental sensors 20 times in an anomalous manner, sending the robot to investigate anomalies five times
each in four locations (a bedroom, hallway, kitchen, and bathroom); the robot asked if it should call for help, to which the experimenter answered affirmatively half of the time and otherwise negatively. The average time required for the robot to arrive at a location and average accuracy in recognizing a human's response correctly were computed.

\subsection{Results}

The prototype required an average of 13.8s (SD: 7.9) to go to the bedroom, hallway, and kitchen. (The data for the bathroom were not used because the robot's initial position was near the bathroom and it only had to turn.) 
Verbal responses from the experimenter at the anomaly location were correctly recognized 76.9\% of the time, with problems arising due to the timing of when the robot should recognize.

Thus, results were mostly successful but indicated room for improvement. Open areas in a single floor apartment could be reached in under a minute (although real-world problems of stairs and blocked passages were not addressed) and a robot could acquire other information in addition to a verbal response to determine if a person is in danger (for example, visual detection of lip movement could confirm when a robot should seek to recognize, and pose detection could indicate if a fall had occurred).

\section{Prototype 8 Useful application (fall emergency): Health assessment}

Asking if a human is okay when simple sensors detect an anomaly might not always work, e.g., if sensor coverage is incomplete (e.g., in the presence of occlusions) or a person is unresponsive.
We propose that it would be helpful if a home robot would also be able to autonomously estimate the health states of detected people, possibly while patrolling.
Some teleoperated robots have been designed to facilitate first aid--e.g., to bring a defibrillator to a specified location [Katz 2015], or to remotely observe and conduct surgery on a soldier on a stretcher [Martinic 2014]--but medical staff might be far away or busy and teleoperation can require effort and skill. 
We built a prototype capable of autonomously assessing health of fallen persons based on some first aid guidelines, as shown in Figure~\ref{prototype8}, by adding some simple sensors and software to an available platform, and focusing on three steps: detecting emergencies, localizing body parts of interest for first aid, and assessing some vital signs.

\includegraphics[width=\linewidth]{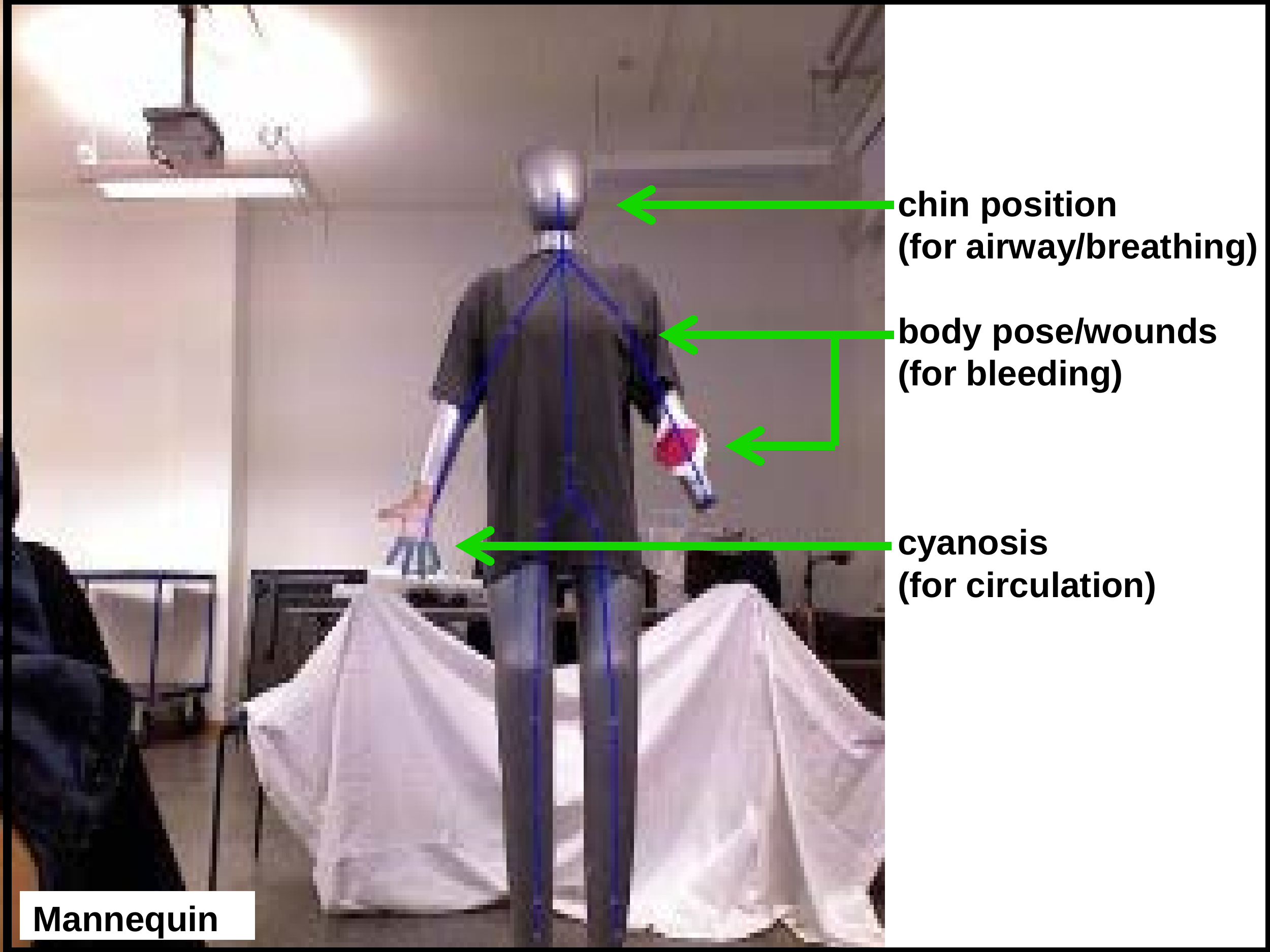}
\begin{figure}[h]
\caption{{\bf Health assessing prototype.}}
\label{prototype8}
\end{figure}

\subsection{Approach}

We wanted our design to leverage typical robotic qualities (using the robot's ability to sense with various sensors and move throughout a home) and produce some human-understandable output. To infer if an emergency occurred, falls and fallen persons were detected: falls were detected by comparing shoulder height displacement of a frontally located person with a threshold, while also noting fall direction, which we expected could be valuable for estimating injury locations. Fallen persons were detected as human-sized, human-temperature anomalies, by comparing the size and temperatures of clustered laser scans within a known map with thresholds during patrolling. The prototype estimated the location of body parts of interest for first aid (chest, hands, chin, mouth, and nose) based on face and skin detection and a simplified prior model. 
To detect faces, which might not be initially visible, the prototype scanned over the anomaly with a camera attached to its arm, and navigated to the other side of the anomaly, while rotating image data, as face angles cannot be known a priori. The estimated pose was visualized over a map of the environment (additionally a visual servoing algorithm was built for the robot to indicate points with a laser pointer).
Furthermore, the prototype checked for relative blueness in the distal portion of the hands to assess circulatory state (peripheral cyanosis), chin pose for airway, speed and normalcy of sound for breathing, and location and rate of expansion of red color for bleeding. 

\subsection{Evaluation}
To assess health, all three steps must be achieved: vital sign estimation relies on recognizing where to measure, which in turn relies on recognizing if there is something to investigate. These steps were evaluated individually as the number of test cases would be prohibitively large for a holistic evaluation (recognition for first aid is a highly
complex task requiring numerous capabilities, each of which must deal with various typical cases). 
For emergency detection, the prototype's ability to detect fall directions and fallen persons was assessed in 40 trials each (20 each for detecting human-sized anomalies and human-temperatures); for falls, a mannequin was pushed forward, backward, or to the sides. For fallen persons cases intended to be confusing for the classifier were used with various poses, sizes, and temperatures. 
For body part localization, face detection success rate and error for chest, hands, chin, mouth, and nose were assessed via 20 and 5 trials. 
For vital signs, cyanosis, chin pose, breathing, and
bleeding were assessed: with 40 samples of six images each with bluing in six regions, and four images with no bluing; 40 samples with chin up or down and face oriented front, side, or downward; 40 samples with ten for regular breathing and 30 for abnormal breathing which was fast, slow, or agonal (resembling gasping sounds emitted near death); 36 samples for location, with six each for head, body, and each limb, and 18 samples for speed, for massive, slight, or no bleeding.

\subsection{Results}
Average accuracy over all parts was 78\%. Accuracy was 85\% for emergency detection: 80\% for detecting fall direction, and 90\% for detecting fallen persons (85\% for anomalies, and 95\% for detecting human temperature). Faces could be detected in 70\% of test cases, for which average error was 0.015m. Average accuracy for vital signs was 79\%: 65\% for cyanosis, 75\% for chin pose, 85\% for breathing, and 91\% for bleeding (97\% location/85\% speed).
We believe that the results are promising for a first prototype. Accuracies were imperfect due to various challenging factors: erroneous estimation of the 3D position of a person's shoulders at close distance in narrow spaces for fall detection, uncertainty of wall positions due to sensor noise for anomaly detection (objects were near walls and
anomaly size was small when the robot could only see the width and not the length of the human body), a cooling hot object for detecting human temperature, extremely angled faces for face and chin pose detection, the simplified model for estimating chest and hand location, low resolution on hand data for cyanosis, noise and high variance for
breathing sounds, closeness of joints for bleeding location detection, and flow rates close to the threshold for bleeding rate detection. If such problems can be mitigated, we expect that robots in homes will make a useful contribution not to replacing human experts, but rather to helping to detect problems and causes quickly.

\section{Prototype 9 Useful application (fall emergency): Finding help}

In a health emergency in a home with more than one person, care facility or public place, aside from calling medical services and checking if a person is okay, a robot can also actively move to try to find a nearby human capable of helping (conducting first aid or driving the victim to a hospital). The challenge is that every second can count: the possibility of a victim being helped should be maximized.
We formulated the problem as follows: nearby humans can be regarded as ``nodes'' on a graph which the robot can visit; each node has a travel cost and reward (expected helpfulness); a time limit is likely (e.g., battery duration, or approx. 5 minutes for cardiac arrest); the robot cannot visit all nodes (all humans); nodes move and can appear or disappear from view; and the robot may pass through a node more than once and might not need to return to its starting point.
Thus, at the high-level, the problem can be structured as a variant of the Traveling Salesman Problem (TSP), e.g. with profits and partially observed [Kataoka and Morito2013]; at the low level, a path planning algorithm such as an A* variant can be used for the robot to move to a node [Hart 1968]. The unique problem was estimating helpfulness, which has been conducted for online reviews [Ghose and Ipeirotis 2011], but not for people in an emergency context.
We programmed a flying robot prototype to approach potentially helpful people in its vicinity, as shown in Figure~\ref{prototype9}.

\includegraphics[width=\linewidth]{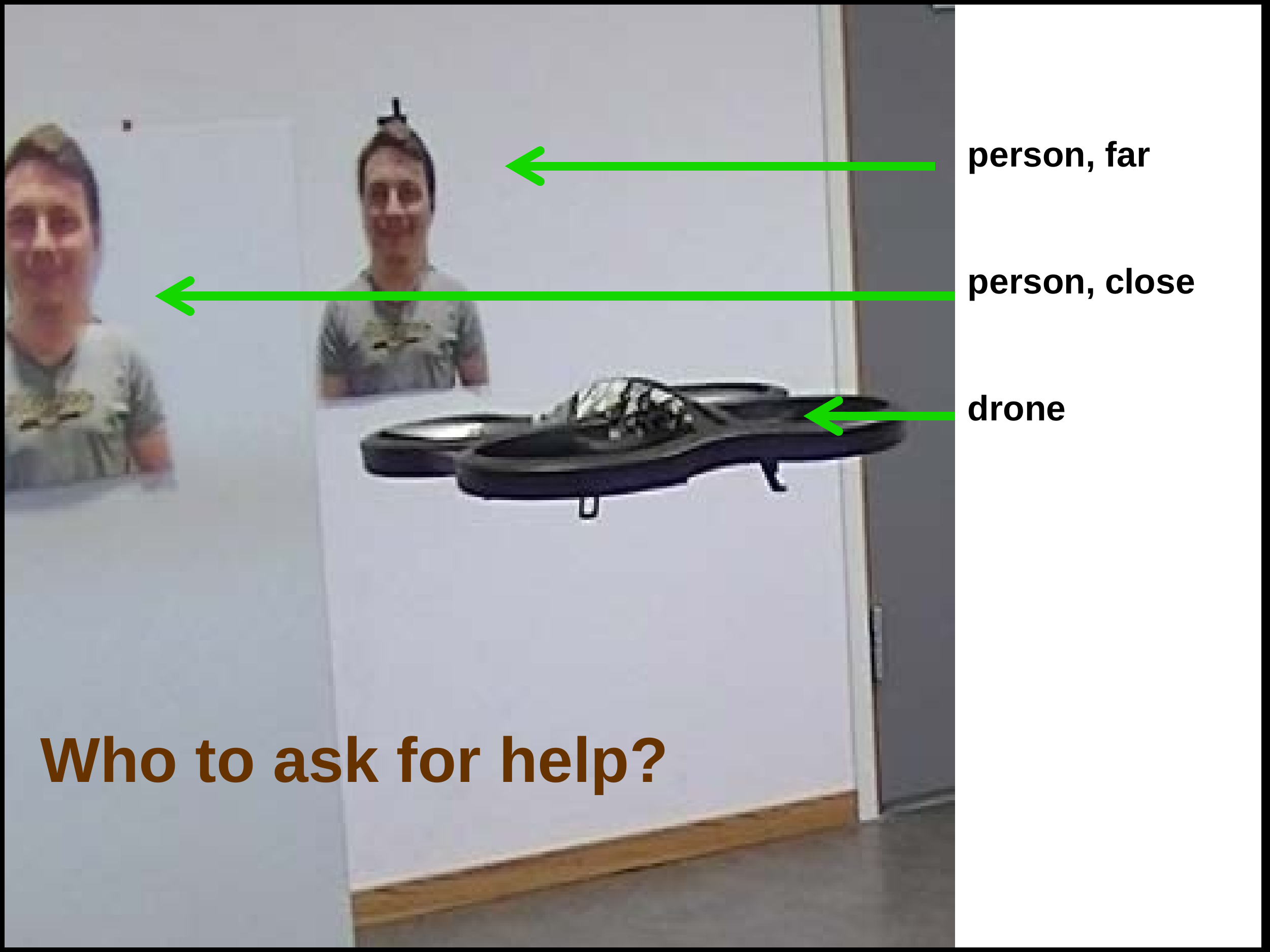}
\begin{figure}[h]
\caption{{\bf Help finding prototype.}}
\label{prototype9}
\end{figure}

\subsection{Approach}

The design required capability for estimating helpfulness and simplified navigation. Helpfulness could be estimated based on various factors, including age, distance, profession, motion, and possession of useful resources. For example, adults, especially medical workers and life guards, who are close by or approaching and possess a cellular phone could be helpful. Conversely young children and elderly might not be physically capable of providing first aid, and valuable time could be lost approaching people far away or moving quickly away from the robot. Thus we designed our prototype to estimate helpfulness based on two rules of thumb: estimated distance based on the size of detected faces, and the height of detected faces (adults are expected to be more helpful and taller than young children). To detect people the prototype used the Viola-Jones approach [Viola and Jones 2001] to detect faces, and heads and shoulders. 

Simplified navigation was conducted via visual servoing; the face of a target helpful person was used as a landmark, with the prototype approaching while turning so as to keep the person's face at the center of its view, until estimated closeness exceeded a threshold. A flying robot was chosen based on the idea that the robot could fly over obstacles, and that a small drone could be carried by a person or installed in environments with elderly.
For a first investigation we also did not consider other problems such as how the robot should communicate, occlusions, difficult poses, or wind.

\subsection{Evaluation}

To evaluate if the prototype could approach persons estimated to be helpful, we placed it in front of two approximately life-sized upper body photos in a room with the door closed and sent a command to find a person to go to for help. Identical photos of one person were used for convenience, and to avoid any confusion which might result due to personal differences in face width, as face size was used to estimate distance.
Two conditions were investigated, distance and height, with 20 trials conducted in total.
In the distance condition (10 trials), the robot was placed 2m and 1.5m away from one photo (five trials each), with the other photo placed farther at 3m for all cases; height for both photos was equal (1.2m).
For height, the robot was placed 2m away and the photos were placed with a vertical distance of approx. 0.3m or approx. 0.15m around 1.2m (five trials each). Positions were chosen such that both photos were in camera view and could be positively detected and a difference could be perceived in distance or height.
During data acquisition successful trials were counted. To be successful, the robot had to approach the closer and higher photos.

\subsection{Results}

The robot approached the correct target 90\% of the time (18/20 trials) but crashed into the wall and a target one time each.
The success rate was reasonable, due to the simplified context. The two failures resulted from some drifting for which the drone's in-built stabilization algorithm could not compensate; this sometimes interfered with detection and we believe was affected by various factors including erratic air currents from the drone's own propellers, low ground contrast, and minor imbalances in propeller alignment or the protective outer hull. We think these problems will be less important in the future as technologies improve.

\section{Discussion}

Nine medium fidelity prototypes were built to explore novel capabilities related to interactiveness and intelligence which might contribute to acceptance of home robots. Some things we learned, some unexpected observations, and some next steps are as follows:


\paragraph{*1 Energy harvesting}
We learned that it is possible to leverage thermal, kinetic, and optic energy harvesting for a small held social robot to provide visible reactions in typical conditions (approx. 20$^{\circ}$C, $>$1300Pa, $>$300 lux).
A nice but unexpected corollary to using freely balanced ears was that the prototype reacted to being picked up or shaken (proprioceptive behavior).
Future work will involve combining low energy parts (microcontroller, sensors, and wireless) to transfer important human vital signals such as body temperature, pulse, respiratory rate, and oximetry to a computer for processing (for example temperature could be measured at low current with a SMT172 sensor and transmitted wirelessly with Silicon Labs EZR32).

\paragraph{*2 Private breath sensing}
We learned that for a close interaction with a small held social robot a breath sensor can be used to estimate human position in an unobtrusive way with an average reaction time of six seconds.
A nice but unexpected insight was that the prototype would react not only when facing away or behind an occlusion (which we had expected), but also to speech.
Future work will involve improving performance by fusing data from multiple sensors.

\paragraph{*3 Fun reaching motions}
We learned that a good and playful impression could be elicited over several minutes by blending a straight functional motion with curved playful components and following some simple heuristics.
One unexpected observation was that a reversed u-shaped relation was observed between perceived playfulness and motion length, suggesting that such motions should be neither too short nor too long.
Future work will investigate playfulness in other kinds of motion such as locomotion or with objects and how other desirable characteristics can be evoked, such as cuteness or coolness.

\paragraph{*4 Adaptive size-changing motions}
We learned that our participants typically perceived expansions as expressing threat or incredulousness, and contractions as expressing fearfulness, attentiveness, and attractiveness.
Unexpected was that expansions in both height and width seemed unnatural due to dispersion of the face, and widening seemed happy due to expansion of the mouth.
Future work will generate more stable structures which can expand in various ways (due to drooping our prototype was placed on the floor in front of interacting people), and test local expansions (e.g., skew motions, or only the robot's body and not the eyes or mouth).

\paragraph{*5 Self-fixing appearance}
We learned that a robot could seek to fix a simple flaw in its own appearance by navigating to detect its own reflection with a success rate of 76.9\% in approx. 150 seconds, discovering anomalies 71\%, and modifying itself.
Unexpected for us were some of the false positives detected by our prototype, which looked little like the robot to us (we believe this was a result of our prototype not being able to deal well with varying illumination and variations in distance and orientation to the mirror, and indicates that improvement will be possible).
Future work will include dealing with various kinds of anomalies (e.g., additive, aligning) and accessing a larger area of the robot for fixing.

\paragraph{*6 Transparent appearance}
We learned that our participants typically perceived transparification as the robot becoming dormant, and that care should be given when used as a reaction to avoid an impression of malfunctioning. 
This was unexpected to us, as we had predicted it would be perceived as a change in valence, not arousal; furthermore, playfulness was not perceived (possibly also because positive valence was not communicated).
Future work will explore local changes, and partial changes, and other mechanisms for transparency, also for other components.

\paragraph{*7 Going to victim}
We learned that a robot, based on processing data from some simple environmental sensors, can reach various locations fairly quickly (13s in a small space, which we estimate to be several minutes for a one-floor Swedish home of average size), and interact via speech with a success rate of 76.9\% to estimate if a human
wanted the robot to call for help. 
Unexpected was that problems occurred from the robot hearing itself speak or not hearing a fast response from a human. 
Future work will involve using visual cues to better tell when a human is speaking, and detecting confusion (e.g., via Glasgow Coma Scale, when a human's assessment of whether help is required may not be correct, as in the case of some stroke victims).

\paragraph{*8 Health assessment}
 We learned that a prototype could be constructed to assess a person's health state in emergencies in a simplified context, based on some first aid guidelines and some simple sensors, with an accuracy of 78\%. 
Unexpected was that the remote temperature sensor was useful even when clothes were worn and at distances of several meters. 
Future work will involve improving performance (e.g., by learning models for a specific person), and using localized body parts and vital sign assessment to try to perform some simple first aid actions on a mannequin (first aid on a human can result in injuries such as broken ribs; therefore development should first aim to demonstrate reliable, safe performance on a mock-up).

\paragraph{*9 Finding help}

We learned that a prototype could be constructed to make a decision about people's helpfulness and approach a nearby target with a success rate of 90\% in a simplified context. 
Unexpected was that even in a closed room, drift was a large problem. 
Future work will involve trials in more complex conditions (e.g., outdoors in the presence of wind), also by using TSP/A* approaches; better estimation of helpfulness by detecting other cues; and finding a good strategy for communicating information about an emergency.
\paragraph{}

In general, we believe that there is some overlap in the knowledge obtained from creating prototypes. For example, the reaching prototype was used to investigate how to generate fun behavior, but some insight can also be drawn from other prototypes: the freely swinging ears on the energy harvesting prototype were fun because they provided quick fine feedback; breathing on a prototype seemed enjoyable possibly because it was anomalous (we usually don't breath on people or pets); and impressions of playfulness in the size-changing prototype, but not in the transparent prototype, could suggest the importance of dynamic motions for generating fun visual behavior.

It should be noted that the current results are limited by the highly exploratory nature of the work. Further studies with high-fidelity prototypes and more participants will be conducted. Also, in cases where our focus was primarily on the robot functionality, such as energy harvesting, we will also evaluate interactions. 
Furthermore, we should investigate if improved services will result from combining capabilities into a single prototype or allowing prototypes to communicate with each other (for example we have built a transparent structure which can become larger or smaller).

Despite the limitations, we think the current results could provide some insights for other applications: as one example, for autonomous vehicles, which we believe are closely related to robots and might also affect people in their everyday lives.
Like robots, vehicles can also make use of thermal, kinetic, and solar energy; e.g., vibration and moving parts such as wheels, as well as passengers' body heat and motions on the wheel and gas pedal could be used to power a phone which could be used in emergencies even if the car battery has a problem. 
Breath sensors can also be used to detect alcohol levels or carbon monoxide. Enjoyable motions of actuators (such as the windshield wipers) could entertain children. 
Adaptive size-changing motions can allow vehicles to expand for stability or contract for compactness when parking or on narrow roads or passing under overpasses. 
Self-fixing will also be a probable next step as autonomous self-diagnosis becomes more advanced.
Transparency has already been incorporated in a simplified form into a concept car via LEDs, and ``heat transparency'' has been incorporated into some tanks; possible benefits will be privacy, especially in unsafe areas, as well as improved situation awareness (blind spots can be seen).
In emergencies, speech can be used to wake a driver who has fallen asleep or to keep alert a sleepy driver, or a robot could be sent from a station to find a person and check that they are alright. 
Speech could also help with drivers who have partial blindness, by describing driving situations when care is required. 
Health assessment could be vital to determine if a driver is unconscious, in danger and may not know it, or otherwise unable to drive (e.g., in the case of stroke or epilepsy); in such cases the vehicle can safely stop or drive to a hospital. 
If hospitals are far away, finding help could also be useful, although constraints will be different than a flying robot (an algorithm for a car should drive on roads). 
Thus, in general, new interactive capabilities could provide customer value in various ways, including enhancing mobility and equality in the transportation system (people who are capable of driving should not be prohibited from driving only because they have a medical condition which might require occasional assistance).

In conclusion,  to investigate some new interactive capabilities for home robots which could facilitate acceptance, we used a mid-fidelity prototyping approach which allowed us to acquire some basic insights while avoiding large costs in time and effort.
Throughout the current work, we observed high complexity both in the problem area as a whole, where many possible goals presented themselves, and in each sub-area, where for each different prototype various assumptions had to made.
Using a mid-fidelity prototyping approach allowed us to quickly redraw some of our expectations and assumptions, focus on specific questions we had, and acquire a general lay-of-the-land from various perspectives.
In this sense we feel that our results support a recent prescription for user experience design in (social) HRI, which advocated the importance of iterative design, properly defined goals (due to the non-triviality of designing positive experiences), and usage of a variety of evaluation methods to avoid bias in the resulting knowledge [Alenljung et al., to appear].
We also feel that this type of prototyping is important for addressing user experience issues early on and in an integrated manner - rather than later on, when in many cases it might be too late, or in an isolated fashion, which might be less informative. Thus, we believe that the results of the current study, in addition to providing some information for designers of home robots, also generally suggest the usefulness of a mid-fidelity prototyping approach for social HRI.

\section{Acknowledgments}
We thank the volunteers who participated in the experiment, and everyone else who helped. This work used the Halmstad Intelligent Home.

\bibliographystyle{ACM-Reference-Format-Journals}


\end{document}